\documentclass[sigconf,nonacm]{acmart}

\usepackage{amsmath}

\usepackage{amssymb}
\usepackage{booktabs}
\usepackage{graphicx}
\usepackage{xcolor}
\usepackage{float}
\usepackage{placeins}
\usepackage{multirow}
\usepackage{longtable}
\usepackage{array}
\usepackage{calc}
\usepackage{enumitem}
\usepackage{tablefootnote}
\providecommand{\tightlist}{%
  \setlength{\itemsep}{0pt}\setlength{\parskip}{0pt}}

\makeatletter
\@namedef{thenone}{}
\makeatother

\renewcommand\footnotetextcopyrightpermission[1]{}
\begin{document}

\title{AliyunConsoleAgent: Training Web Agents in Real-World Cloud Environments via Distillation and Reinforcement Learning}

\author{Bojie Rong\textsuperscript{*\dag}, Zheyu Shen\textsuperscript{*\dag}, Qiaoping Wang, Pengfei Kang, Yang Xu, Yawen Wei, Hanyu Wu, Zhi Zhao, Leihao Pei, Linquan Jiang}
\affiliation{\institution{Alibaba Cloud}\country{China}}

\thanks{\textsuperscript{*}Equal contribution. \textsuperscript{\dag}Corresponding authors: \texttt{rongbojie.rbj@alibaba-inc.com}, \texttt{shenzheyu.szy@alibaba-inc.com}.}

\begin{abstract}
We present AliyunConsoleAgent, a web agent framework for automated documentation verification in real-world cloud consoles. Major cloud platforms encompass hundreds of products with rapid feature iteration, causing console UIs to frequently diverge from their corresponding documentation. Verifying that documented procedures accurately reflect the current console and can be executed end-to-end demands an estimated 4 million recurring inspections annually, yet manual coverage remains below 1\%. While agent systems built on frontier proprietary models achieve high success rates, their prohibitive cost and data privacy constraints preclude large-scale deployment. We propose a two-stage training paradigm: supervised fine-tuning (SFT) on distilled frontier-model trajectories, followed by reinforcement learning using Group Relative Policy Optimization (GRPO) and a dual-channel outcome reward model in real cloud environments. To support large-scale RL training, we construct a high-determinism rollout system featuring Terraform-based resource pre-provisioning and LLM-driven on-demand provisioning, which effectively isolates environment noise from the training signal. We further introduce a rule-based reward evaluation protocol grounded in backend audit logs, providing objective, reward-hacking-resistant outcome judgment. Our model evolves from mechanical instruction following to autonomous decision-making with cloud console and product-specific understanding. Experiments on a challenging 278-task benchmark where the best frontier model achieves only 65.34\% demonstrate that AliyunConsoleAgent-32B achieves a 63.52\% mean success rate---a 20.24 percentage-point improvement over the base model, narrowing the gap to the best frontier proprietary model to 1.82 pp (bootstrap 95\% CI $[-1.27, 7.39]$)---at 92\% lower inference cost.
\end{abstract}


\maketitle
\renewcommand{\shortauthors}{Rong and Shen, et al.}

\section{Introduction}\label{introduction}

\begin{figure}[t]
\centering
\includegraphics[width=0.96\columnwidth]{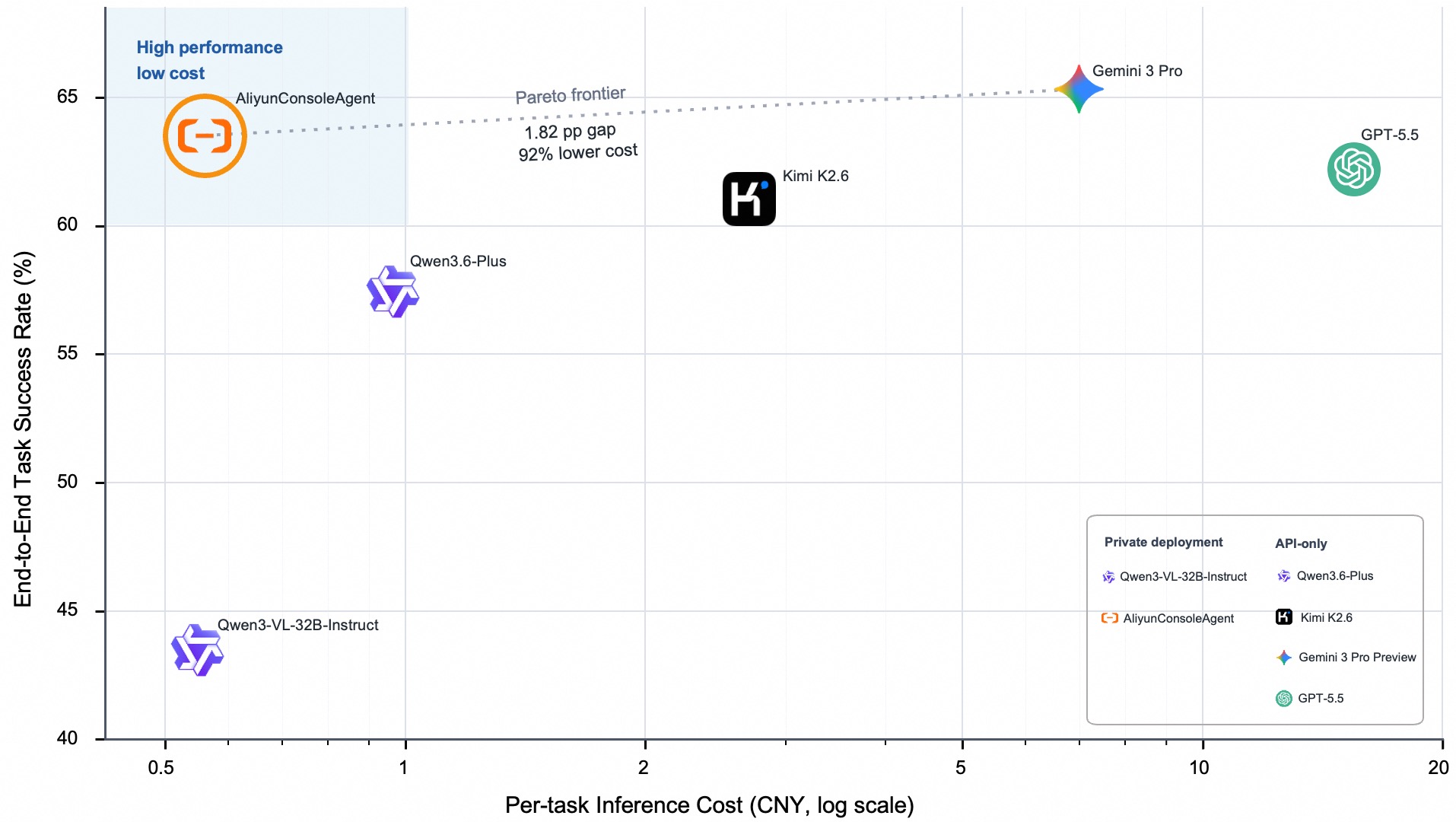}
\caption{Per-task inference cost vs.\ pass@1 success rate. Blue-shaded region marks high-performance, low-cost models under private deployment. AliyunConsoleAgent-32B (SFT+GRPO) narrows the gap to Gemini 3 Pro Preview to 1.82 pp at 92\% lower cost.}
\label{fig:cost-success}
\end{figure}

\begin{figure*}[t]
\centering
\includegraphics[width=0.62\textwidth]{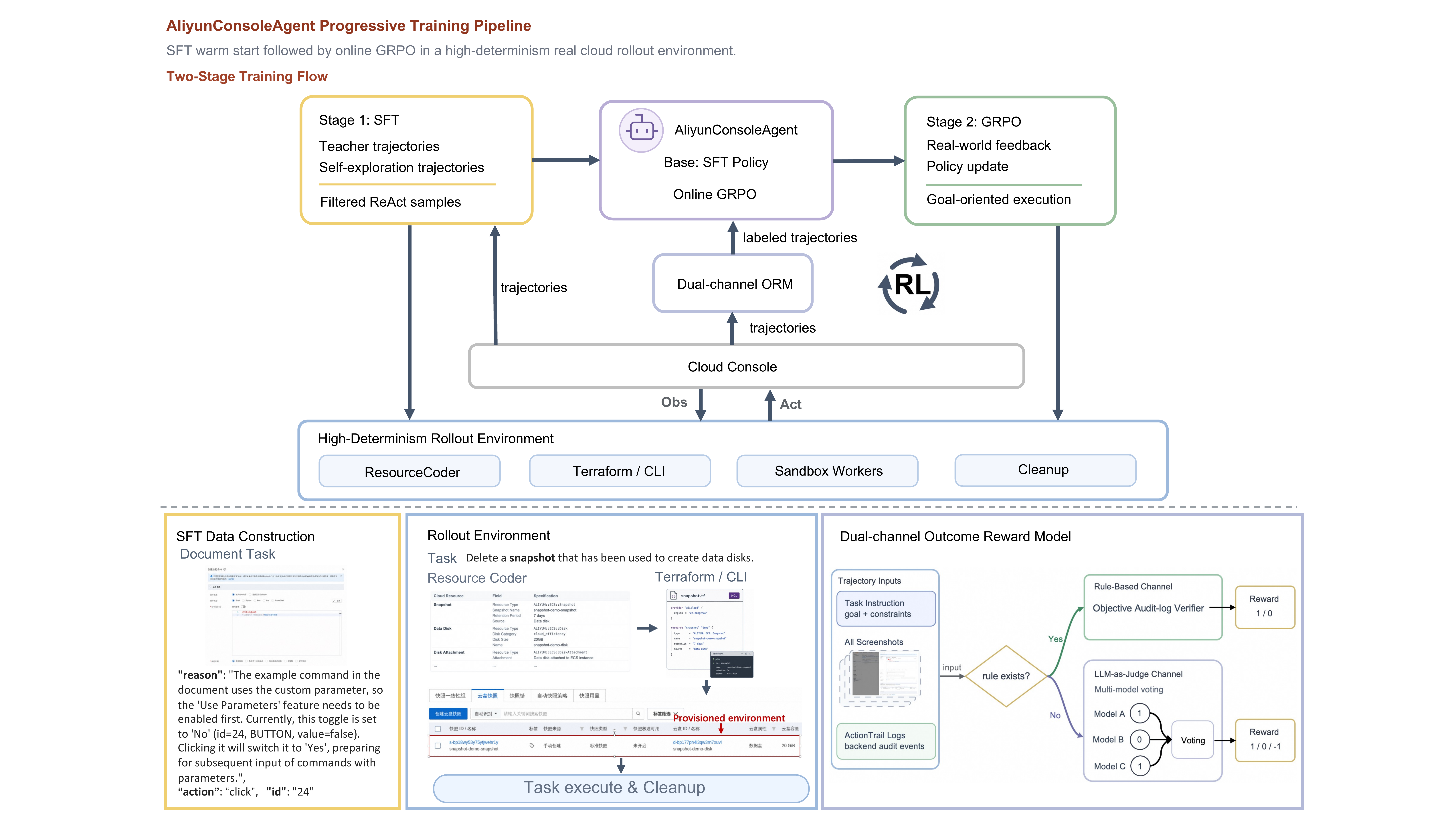}
\caption{AliyunConsoleAgent training pipeline: two-stage SFT+GRPO training atop the high-determinism Rollout Environment, with SFT data construction, rollout execution/cleanup, and dual-channel ORM.}
\label{fig:training-pipeline}
\end{figure*}

Cloud platforms face a structural \textbf{documentation drift} problem: console UIs evolve under fast release cycles, while their accompanying documentation is authored and maintained manually. With no automated synchronization between the two, documentation gradually diverges from the live console, leaving users to follow procedures whose steps no longer match the current interface. Detecting such inconsistencies at scale is infeasible by hand: by estimate, the workload requires 4 million recurring inspections annually to keep documentation current, each inspection demanding approximately 1.6 hours of manual effort, while manual coverage today remains below 1\%. This gap makes automated detection an urgent industry need. \textbf{Execution-as-Verification} addresses this by driving an agent to faithfully execute each documented step end-to-end.

We have deployed an initial production system using frontier proprietary models, confirming the value of automated detection. However, scaling to full coverage and periodic detection faces \textbf{cost and privacy bottlenecks}: proprietary models incur prohibitive per-call costs, and transmitting contexts containing sensitive cloud metadata to external services poses compliance risks. Training a lightweight 32B model with on-premise deployment support becomes the necessary path. Following the SFT+RL paradigm established by recent GUI agents \cite{qin2025uitars,bytedance2025uitars2}, the key challenge lies in constructing a stable and consistent RL training environment. Cloud console tasks have strong \textbf{resource dependencies}: missing prerequisites (e.g., a specific type of ECS (Elastic Compute Service) instance) produce environmental failures during exploration that are indistinguishable from genuine agent errors, polluting the reward signal and preventing policy convergence.

To address these challenges, we propose the AliyunConsoleAgent training framework (Figure~\ref{fig:training-pipeline}). Our main contributions are:
\begin{itemize}
\tightlist
\item \textbf{High-Determinism Rollout Environment with Audit-Log Verification:} We build a real-world cloud-console rollout environment with Terraform-based resource pre-provisioning and ResourceCoder on-demand provisioning, together with ActionTrail (Alibaba Cloud's API audit log service) verification for objective, reward-hacking-resistant training and evaluation.
\item \textbf{Cost-Effective Private Agent Training:} We propose an SFT-then-GRPO training paradigm with a dual-channel Outcome Reward Model. The resulting AliyunConsoleAgent-32B achieves a 63.52\% mean success rate on a 278-task benchmark---approaching Gemini 3 Pro Preview (bootstrap 95\% CI includes zero, $p > 0.05$)---at 92\% lower cost.
\item \textbf{Production-Scale Cloud Documentation Verification:} We deployed a verification framework using frontier models to audit 54,000+ documented procedures, identifying 4,399 confirmed defects accepted by product teams. We demonstrate that our trained AliyunConsoleAgent achieves the performance necessary to serve as a privacy-compliant, drop-in replacement, projecting a 92\% inference-cost reduction for sustainable full-scale deployment.
\end{itemize}

\noindent To ensure reproducibility, we open-source our evaluation benchmark, filtered training data, Rollout environment infrastructure, and model training code at \url{https://github.com/AlibabaResearch/aliyun-console-agent}.

\section{Related Work}\label{related-work}

\textbf{GUI and Web Agents.}
Building autonomous agents that interact with graphical user interfaces is an active research area. Early benchmarks (World of Bits \cite{shi2017world}, Mind2Web \cite{deng2024mind2web}, WebLINX \cite{lu2024weblinx}) progressively scale web-agent evaluation toward real-world interaction, while vision-centric approaches such as UI-TARS \cite{qin2025uitars} and SeeClick \cite{cheng2024seeclick} pioneer end-to-end visual understanding and element grounding from screenshots. Standardized benchmarks like WebArena \cite{zhou2023webarena}, OSWorld \cite{xie2024osworld}, and BrowserGym \cite{drouin2024browsergym} cover web, desktop, and enterprise environments, and EvoCUA \cite{xue2026evocua} introduces scalable synthetic experience for evolving computer-use agents. These works primarily target open-domain tasks in sandboxed or semi-controlled settings; real-world cloud console operations present unique challenges---extreme UI density with hundreds of interactive elements per page, strong resource state dependencies, and non-deterministic backend behavior---creating a significant domain gap.

\textbf{RL for Agent Training.}
Reinforcement learning has emerged as a key technique for improving agent autonomy beyond supervised imitation. Foundational RL methods---PPO \cite{schulman2017ppo} and DPO \cite{rafailov2023dpo}---underpin recent agent training pipelines, with DigiRL \cite{bai2024digirl}, WebRL \cite{qi2024webrl}, and UI-TARS-2 \cite{bytedance2025uitars2} applying online RL to simulated and GUI environments. Mobile extensions (MobileRL \cite{xu2025mobilerl}, MAI-UI \cite{zhou2025maiui}) and self-evolving approaches (SEAgent \cite{sun2025seagent}, AgentQ \cite{putta2024agentq}, ZeroGUI \cite{yang2025zerogui}, AutoGLM \cite{liu2024autoglm}) further explore autonomous learning in real environments, while DeepSeek-R1 \cite{deepseek2025r1} demonstrates GRPO's scalability to large reasoning models. Our work differs in two key aspects: (1) we conduct RL in a real production cloud environment rather than sandboxed simulators, requiring explicit environmental noise isolation through our Rollout infrastructure; and (2) we address the unique challenge of resource-dependent state management, where environmental failures (rather than agent errors) dominate the failure signal without proper provisioning.

\textbf{Multimodal Agents and Knowledge Distillation.}
The ReAct paradigm \cite{yao2023react} establishes interleaved reasoning and acting as the standard decision-making framework for LLM-based agents, which we adopt for structuring our SFT training data. Set-of-Mark (SoM) visual prompting \cite{yang2023setofmark} enables precise GUI element identification by overlaying numeric labels on screenshots, serving as our primary observation modality. For training data construction, knowledge distillation from frontier models has proven effective---using capable teacher models to generate high-quality trajectories that are then filtered and used to fine-tune smaller student models \cite{furuta2024multimodal,he2024webvoyager}---and OS-Genesis \cite{sun2024osgenesis} further automates trajectory construction via reverse task synthesis. Our base model builds on the Qwen3-VL architecture \cite{bai2025qwen3vl}, which provides strong vision-language understanding as a foundation for domain-specific fine-tuning. We extend these techniques to the cloud console domain, combining trajectory distillation with real-world RL to handle the unique environmental stochasticity of production cloud platforms.

\section{Methodology}\label{methodology}

\subsection{Problem Formulation}\label{problem-formulation}

We formalize the agent-console interaction as a POMDP. The state $S$ encompasses both the visible frontend UI and hidden backend resource configurations (e.g., VPC networks, IAM privileges). Because these hidden configurations cause identical GUI actions to yield different outcomes, we introduce an on-demand provisioning action $a_{env}$ to stabilize transition dynamics. The full action space $\mathcal{A} = \mathcal{A}_{gui} \cup \{ a_{env}, \texttt{task\_complete}, \texttt{task\_unachievable} \}$ combines GUI interactions, $a_{env}$, and terminal signals. At each step, the agent receives an observation $z_t = (I_t, E_t, H_t)$: a SoM-annotated screenshot \cite{yang2023setofmark} ($I_t$), extracted DOM text ($E_t$), and action history ($H_t$). The reward $R$ combines rule-based ActionTrail verification and LLM-based outcome evaluation (Section~\ref{grpo-phase}).

\subsection{WebAgent Framework}\label{webagent-framework}

\begin{figure*}[t]
\centering
\includegraphics[width=0.62\textwidth]{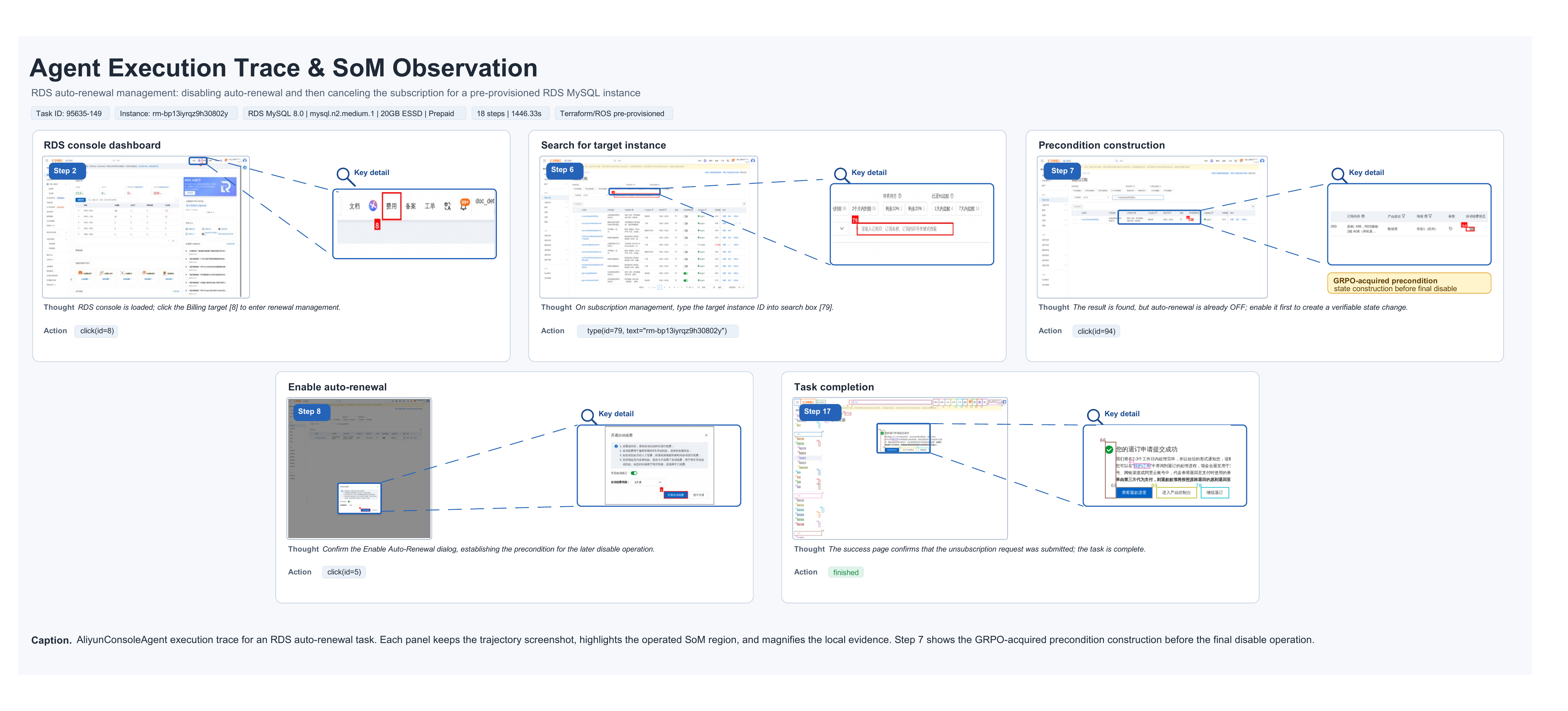}
\caption{Execution trace for an RDS auto-renewal task. The GRPO model autonomously enables auto-renewal first to create the precondition for a verifiable ``disable'' operation.}
\label{fig:execution-trace}
\end{figure*}

The agent operates in a continuous Observation-Reasoning-Action loop following the ReAct paradigm \cite{yao2023react}. At each step, it captures a full-page screenshot with SoM annotations\footnote{We adopt SoM-based id targeting over direct coordinate prediction because cloud console pages typically contain hundreds of small interactive elements, substantially denser than typical web pages or desktop apps, and our internal evaluations show that current vision-language models produce noticeably less accurate coordinates than id selections in this regime.} and extracts DOM text elements to form the observation $z_t$. The agent produces an explicit \texttt{Thought} trace, assessing task progress, analyzing the current UI state, and identifying the next sub-goal, before selecting one of four kinds of actions: a GUI interaction $a_t \in \mathcal{A}_{gui}$ executed via Playwright on a headless Chromium instance; \texttt{task\_complete} to declare success; \texttt{task\_unachievable} when the procedure cannot proceed in the current environment; or the on-demand provisioning action $a_{env}$, which pauses GUI operations and triggers an external ResourceCoder. ResourceCoder is an LLM-based coding agent that writes, executes, and self-corrects Terraform \cite{hashicorp2024terraform} / Aliyun CLI scripts in a sandboxed loop; on successful provisioning, it returns a structured resource report injected directly into the agent's context. Since a documentation procedure can be decomposed into sub-steps following its ordered-list structure, we assign a per-sub-step interaction budget of 40---effectively a dynamic max-turn limit that scales with task complexity. The task finishes when the final sub-step succeeds, any sub-step exhausts its budget, or the agent emits \texttt{task\_unachievable}.

\section{Rollout Environment}\label{rollout-environment}

\begin{figure*}[t]
\centering
\includegraphics[width=0.62\textwidth]{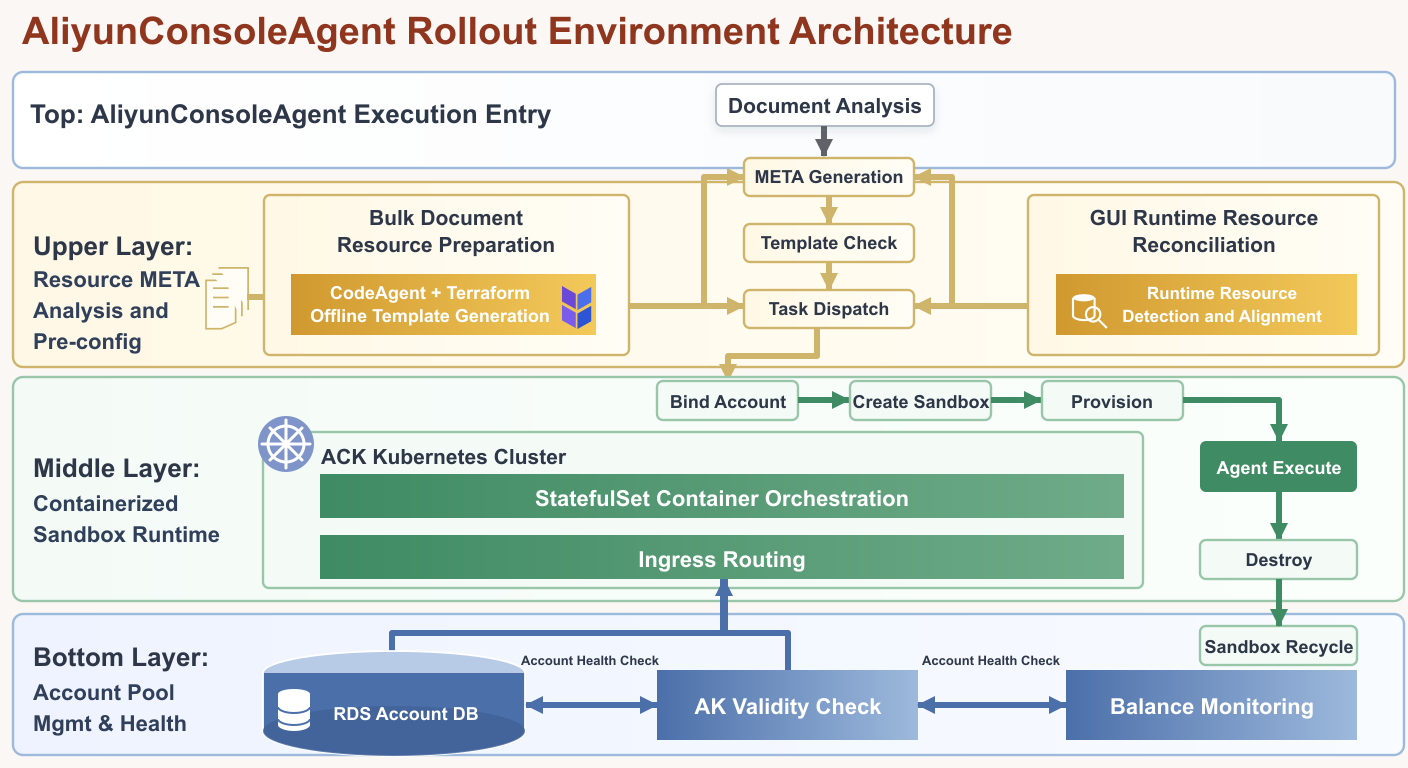}
\caption{Layered architecture of the Rollout environment: Account Pool, Sandbox Execution, Resource META Provisioning, and Agent execution entry.}
\label{fig:rollout-arch}
\end{figure*}

Building a stable Rollout environment is critical for both RL training and evaluation. Unlike open-domain web browsing, cloud console tasks exhibit strong \textbf{resource dependencies}: operations such as ``modify an RDS (Relational Database Service) whitelist'' or ``unsubscribe from an ECS instance'' implicitly require specific backend resources as prerequisites. Without proper provisioning, the agent encounters systematic failures---empty resource lists, pre-existing but incorrectly configured resources leading to UI misoperations, or concurrency conflicts on shared accounts---that are indistinguishable from decision errors. The two naive baselines trade speed for accuracy without resolving this tension: (i)~parallel execution on isolated empty accounts is fast but lacks prerequisites, so most tasks fail before the agent can act; (ii)~sequential execution on a shared account accumulates state from prior tasks but is slow and inherently serialized, making large-scale rollout infeasible. For RL training, this environmental noise is particularly damaging, as ``failure feedback'' originating from resource absence rather than agent errors dilutes the policy gradient and prevents convergence.

To address this, we construct a four-layer Rollout architecture (Figure~\ref{fig:rollout-arch}). The \textbf{Account Pool Management Layer} maintains a pool of isolated test accounts with automated credential rotation and financial health monitoring, ensuring that each concurrent rollout task operates on a dedicated account to prevent cross-task interference. The \textbf{Sandbox Execution Layer} provides containerized environments via Alibaba Cloud Container Service for Kubernetes (ACK) with StatefulSet orchestration for task-level isolation and elastic scaling. The \textbf{Resource META Offline Provisioning Layer} is the core innovation: for each document, we define a Resource META five-tuple $\mathcal{M} = \langle D, K, C_{\text{create}}, C_{\text{verify}}, C_{\text{destroy}} \rangle$, comprising a natural language dependency description ($D$), structured key-attribute configuration ($K$, exemplified in Table~\ref{tab:key-attributes}), and executable provisioning, verification, and destruction code ($C_{\text{create}}, C_{\text{verify}}, C_{\text{destroy}}$). To ensure cross-account portability, all templates create dependency resources (VPCs, vSwitches, security groups) from scratch with randomized naming rather than referencing pre-existing resources. A ResourceCoder-driven pipeline generates these Terraform \cite{hashicorp2024terraform} templates offline through a three-stage process: (1)~dependency identification---analyzing document structure to extract resource constraints and infer implicit dependencies; (2)~template generation with closed-loop correction---iteratively executing templates in a sandbox with automatic error diagnosis and resolution (e.g., substituting discontinued specifications, switching availability zones on quota exhaustion); (3)~verification and destruction code generation. Through this pipeline, we constructed validated META records for 4,000+ documents. At runtime, the system reads the corresponding META record and provisions resources via $C_{\text{create}}$; after task completion, $C_{\text{destroy}}$ restores the account to a clean initial state, enabling reliable reuse across training episodes.

{\def\LTcaptype{none}
\begin{table}[t]
\caption{Example key attribute configuration $K$ in the Resource META for an ECS unsubscription verification task.}
\label{tab:key-attributes}
\small
\begin{tabular}{@{}lll@{}}
\toprule
Attribute & Description & Value \\
\midrule
\texttt{charge\_type} & Billing & \texttt{PrePaid} \\
\texttt{instance\_type} & Spec & \texttt{ecs.n4.large} \\
\texttt{image\_id} & OS image & \texttt{aliyun\_3\_x64\_20G...} \\
\texttt{zone} & Availability Zone & \texttt{cn-beijing-h} \\
\texttt{disk\_category} & Disk type & \texttt{cloud\_efficiency} \\
\bottomrule
\end{tabular}
\end{table}
}

The \textbf{Runtime On-demand Provisioning Layer} complements offline META templates by handling unforeseen resource gaps---implicit dependencies not documented in prerequisites, cascading cross-product dependencies, or specification mismatches caused by platform backend changes. When the agent detects an environmental blockage (e.g., empty resource lists, ``resource does not exist'' prompts), it outputs $a_{env}$, triggering the ResourceCoder to dynamically provision missing resources and inject a structured report back into the agent's context. Successfully converged runtime templates are persisted back to the offline META platform for future reuse, keeping creation and cleanup code paired. Together, the offline META layer covers explicitly documented dependencies while the runtime layer addresses implicit ones, forming a complete ``provision---execute---recover---destroy'' closed loop. Post-task cleanup follows dependency order (higher-level resources before lower-level dependencies), restoring the account to its initial state.

This architecture resolves the speed-accuracy tradeoff of the two naive baselines. On ECS, parallel execution with empty-account resource provisioning lifts success from 33.81\% (baseline (i)) to 84.39\% (+50.58 pp), while shrinking a full doc-tree audit from $\sim$5 days under baseline (ii) to roughly 1 hour given sufficient accounts and machines. A 10-product cross-validation confirms that this throughput gain does not sacrifice end-to-end accuracy: 80\% average success, 4 pp above the sequential baseline (ii). For RL training, the same 50.58 pp gain translates to substantially cleaner reward signals, enabling GRPO to attribute failures to genuine agent decision errors rather than environmental artifacts.

\section{Model Training}\label{model-training}

\subsection{SFT Phase}\label{sft-phase}

The SFT phase equips the base model with foundational cloud console interaction capabilities within our WebAgent framework through large-scale trajectory imitation. A key design decision is enforcing a structured ReAct \cite{yao2023react} decision-making paradigm where the model must produce an explicit \texttt{Thought} trace before each action to assess task progress, analyze the current UI state, and identify the next sub-goal. This structured format ensures that the model develops interpretable reasoning chains that can be inspected and debugged in production. Our dataset is constructed from two complementary sources. First, \textbf{high-quality trajectory distillation}. We collect successful execution traces and apply two-level filtering: task-level, which retains only successful trajectories that complete tasks or find true inconsistencies in documents, and step-level, an LLM-based evaluation to remove ineffective actions and rollout detours. Second, \textbf{autonomous self-exploration}. The model freely interacts with the console from high-level product entry points (e.g., ECS, RDS), autonomously proposing and executing CRUD (create, read, update, delete) tasks to maximize UI coverage across products and interaction types, with successful trajectories filtered at the step level. This self-play mechanism is critical for covering the long tail of UI states and interaction patterns not present in the distilled data. We conduct full-parameter fine-tuning on Qwen3-VL-32B-Instruct \cite{bai2025qwen3vl} using $\sim$160K single-step samples for 1 epoch on 64 GPUs with DeepSpeed ZeRO-2 (learning rate $1.0 \times 10^{-5}$, cosine schedule, BF16 precision).

\subsection{GRPO Phase}\label{grpo-phase}

While SFT provides foundational capabilities, it cannot generalize to the inherent stochasticity of real cloud environments---varying precondition states, asynchronous backend processes, and ambiguous documentation. We employ \textbf{GRPO (Group Relative Policy Optimization)} \cite{shao2024deepseekmath} to train the model through real-world environmental feedback, enabling it to evolve from faithful instruction following to autonomous decision-making. The training pipeline fully decouples rollout execution from gradient computation: a remote Rollout service with dedicated inference endpoints handles all environment interaction and action generation, while the training cluster focuses solely on parameter updates. This separation enables independent scaling of rollout workers for higher data throughput and avoids occupying training GPUs with inference workloads.

\textbf{Outcome Reward Model (ORM).} We design a dual-channel ORM for reliable reward signals. \textbf{Channel 1} (Rule-Based): for tasks with deterministic success criteria, we query ActionTrail audit logs to verify task completion, providing ground-truth binary rewards with zero false positives. \textbf{Channel 2} (LLM-as-Judge Ensemble \cite{zheng2023judging}): for tasks lacking a definitive success-indicating API event in ActionTrail audit logs, we employ an ensemble of two strong general-purpose LLMs as trajectory-level judges. The ensemble requires consensus: only when both judges agree is the reward accepted; disagreements yield an invalid label ($r = -1$) and the trajectory is excluded from training. This directly addresses ORM hallucination and reward hacking \cite{tencent2025cuarewardbench}. On a separate set of 308 expert-labeled trajectories for reward model validation, the ensemble achieves 96.7\% accuracy against human judgments (vs.\ 91.9\% for single-model).

\begin{equation}\label{eq:orm}
r_i = \begin{cases} 1.0 & \text{if ActionTrail confirms success} \\ 1.0 & \text{if both LLM judges agree on success} \\ 0.0 & \text{if both LLM judges agree on failure} \\ -1.0 & \text{otherwise (excluded from training)} \end{cases}
\end{equation}

\textbf{Two-Layer Advantage Normalization.} To handle heterogeneous task difficulties within each training batch, we apply two successive normalization steps. (1) \textbf{Per-task group normalization}: $\hat{A}_{k,i} = (r_{k,i} - \mu_k) / (\sigma_k + \epsilon)$, where $r_{k,i}$ is the binary reward for the $i$-th rollout of task $k$, $\mu_k$ and $\sigma_k$ are the mean and standard deviation of rewards within task $k$'s $G$ rollouts, and $\epsilon$ is a small constant for numerical stability. This eliminates task-level difficulty differences. (2) \textbf{Global cross-task normalization}: $A_{i}^{\text{final}} = (\hat{A}_{i} - \mu_{\text{batch}}) / \max(\sigma_{\text{batch}}, \epsilon)$, where $\mu_{\text{batch}}$ and $\sigma_{\text{batch}}$ are computed over all per-task-normalized advantages in the batch, unifying advantage scales across tasks. When all $G$ rollouts for a task yield identical outcomes, $\sigma_k = 0$ produces zero advantage, effectively skipping uninformative tasks \cite{yu2025dapo}.

\textbf{Policy Update.} Following GRPO \cite{shao2024deepseekmath}, we optimize a clipped surrogate objective with KL regularization. The per-token loss is:
\begin{equation}\label{eq:policy}
\mathcal{L} = -\min\!\bigl(\rho_t \hat{A}_t,\; \text{clip}(\rho_t, 1{-}\epsilon, 1{+}\epsilon)\hat{A}_t\bigr) + \beta \cdot \mathcal{L}_{\text{KL}}
\end{equation}
where $\rho_t = \pi_\theta(a_t|s_t)/\pi_{\text{old}}(a_t|s_t)$ is the importance sampling ratio between the current policy and the policy at rollout time, and $\hat{A}_t$ is the normalized advantage from the previous step. Following ZeroGUI \cite{yang2025zerogui}, we replace the default k3-estimator KL penalty with a per-token MSE form: $\mathcal{L}_{\text{KL}} = \frac{1}{2}(\log \pi_\theta(a_t|s_t) - \log \pi_{\text{ref}}(a_t|s_t))^2$, where $\pi_{\text{ref}}$ is the frozen SFT policy. This provides more stable gradients when the policy diverges significantly from the reference. We set clip range $\epsilon = 0.2$, KL coefficient $\beta = 0.1$, group size $G=16$ with $1.25\times$ oversampling, and learning rate $5 \times 10^{-7}$.

\section{Evaluation}\label{evaluation}

\subsection{Single-Step Evaluation}\label{single-step-evaluation}

Before evaluating end-to-end task completion, we assess atomic action prediction accuracy in a controlled single-step setting. We construct a benchmark of 400 single-step state-action pairs randomly sampled from trajectories across all cloud products. Each sample provides a screenshot with SoM annotations and task instruction; the model must predict the correct next action. Multiple ground-truth actions are annotated per sample to account for valid alternative interaction paths (e.g., clicking a button vs.\ using a keyboard shortcut). To avoid model sampling randomness, we evaluate each model 3 times and average the accuracy for final result. Table~\ref{tab:single-step} summarizes results.

{\def\LTcaptype{none}
\begin{table}[t]
\caption{Single-step accuracy on the 400-sample cloud console benchmark. Models in the ``AliyunConsoleAgent (Ours)'' category share the same SFT pipeline, differing only in base model.}
\label{tab:single-step}
\small
\begin{tabular}{@{}lc@{}}
\toprule
Model & Accuracy \\
\midrule
\multicolumn{2}{@{}l}{\textit{Frontier}} \\
\quad Gemini 3 Pro Preview & 95.25\% \\
\quad GPT-5.5 & 94.75\% \\
\quad Kimi K2.6 & 92.33\% \\
\quad Qwen3.6-Plus & 90.75\% \\
\midrule
\multicolumn{2}{@{}l}{\textit{Open-Source Base}} \\
\quad Qwen3-VL-8B-Instruct & 71.17\% \\
\quad Qwen3-VL-32B-Instruct & 83.00\% \\
\midrule
\multicolumn{2}{@{}l}{\textit{AliyunConsoleAgent (Ours)}} \\
\quad Qwen3-VL-8B (SFT) & 88.84\% \\
\quad \textbf{AliyunConsoleAgent-32B (SFT)} & \textbf{92.75\%} \\
\bottomrule
\end{tabular}
\end{table}
}

After SFT, AliyunConsoleAgent-32B reaches 92.75\% (+9.75 pp over base), exceeding Kimi K2.6 (92.33\%) and Qwen3.6-Plus (90.75\%) and approaching the strongest frontier models (Gemini 3 Pro Preview: 95.25\%, GPT-5.5: 94.75\%) in atomic instruction execution. The SFT model's most common failures appear on under-specified procedures, where documentation gives only a high-level objective or where a step has implicit prerequisites (e.g., an instance must be \texttt{Active} before unsubscription). Without explicit instruction, the SFT model falls back to superficial heuristics (e.g., selecting the first list item), motivating RL to develop the contextual reasoning needed for complex cloud-console tasks.

\subsection{End-to-End Testing in Production}\label{end-to-end-testing}

\textbf{Rule-based Reward for Objective Evaluation.} Unlike screenshot-based or LLM-as-judge evaluation, which can be brittle and introduce model-dependent bias \cite{xue2026evocua,tencent2025cuarewardbench,bytedance2025uitars2}, we introduce a \textbf{Rule-based Reward} protocol that queries Alibaba Cloud ActionTrail audit logs as an objective, binary evaluation signal:
\begin{equation}\label{eq:rule-reward}
R_{\text{rule}}(\tau) = f_{\theta_t}(\mathcal{L}_{a, [t_s, t_e + \Delta]}) \in \{0, 1\},
\end{equation}
where $\tau$ is the rollout trajectory, $\mathcal{L}_{a, [t_s, t_e + \Delta]}$ collects the ActionTrail audit events recorded between task start $t_s$ and task end $t_e$ plus a small buffer $\Delta$ for asynchronous backend events, and $f_{\theta_t}$ is a task-specific verification rule that checks for expected API events and parameter constraints. All verification rules are validated by domain experts to ensure correctness. This design is resistant to reward hacking: the agent must produce verifiable backend state changes rather than merely convincing a judge model.

\textbf{Benchmark Setup.} We sample 278 real cloud product documentation verification tasks across 12 Alibaba Cloud products, comprising 76 standard and 202 hard tasks. \emph{Standard} tasks follow the same distribution as online production workloads, while \emph{Hard} tasks are sampled from flows with high online execution failure rates. All 278 tasks are verified exclusively via rule-based ActionTrail audit rules (no LLM-as-judge in evaluation). The evaluation set is isolated from training data at the document level---no document URL appearing in SFT or RL rollouts is included in the benchmark. The complete evaluation dataset with task specifications and reward functions is publicly available in our GitHub repository. Each task is executed 3 independent times; we report both pass@1 (mean success rate $\pm$ standard deviation across runs) and pass@3 (task succeeds if any run passes), where pass@3 reflects the production setting where failed verifications can be retried. All models share the same Rollout Environment with Terraform pre-provisioning and ResourceCoder dynamic provisioning.

\textbf{Main Results.} Table~\ref{tab:e2e} presents the end-to-end results.

{\def\LTcaptype{none}
\begin{table*}[t]
\caption{End-to-end results on the 278-task benchmark (Rule-based ActionTrail verification). pass@1 reports mean $\pm$ std across 3 runs; pass@3 = task succeeds if any run passes. Cost is in CNY per task.}
\label{tab:e2e}
\small
\begin{tabular}{@{}llccc@{}}
\toprule
Model & Deployment & Cost & pass@1 & pass@3 \\
\midrule
Qwen3-VL-32B-Instruct (Base) & Private & 0.56 & 43.28{\scriptsize$\pm$0.50}\% & 60.79\% \\
AliyunConsoleAgent-32B (SFT) & Private & 0.56 & 56.89{\scriptsize$\pm$0.83}\% & 72.66\% \\
\textbf{AliyunConsoleAgent-32B (SFT+GRPO)} & \textbf{Private} & \textbf{0.56} & \textbf{63.52{\scriptsize$\pm$1.86}\%} & \textbf{75.18\%} \\
\midrule
Qwen3.6-Plus & API-only & 0.97 & 57.69{\scriptsize$\pm$1.50}\% & 72.66\% \\
Kimi K2.6 & API-only & 2.70 & 60.79{\scriptsize$\pm$0.72}\% & 73.38\% \\
Gemini 3 Pro Preview & API-only & 7.00 & 65.34{\scriptsize$\pm$2.15}\% & 79.86\% \\
GPT-5.5 & API-only & 15.50 & 62.08{\scriptsize$\pm$2.00}\% & 76.09\% \\
\bottomrule
\end{tabular}
\end{table*}
}

All AliyunConsoleAgent-32B variants share identical deployment infrastructure (4$\times$ L20 GPUs at 0.4 QPS (queries per second)), yielding a per-task cost of approximately 0.56 CNY. Compared with API-only baselines, AliyunConsoleAgent-32B (SFT+GRPO) outperforms Qwen3.6-Plus (+5.83 pp), Kimi K2.6 (+2.73 pp), and GPT-5.5 (+1.44 pp) while remaining substantially cheaper. Against Gemini 3 Pro Preview ($\sim$7.0 CNY per task), it reaches 63.52\% pass@1 and narrows the gap to merely \textbf{1.82 pp} at \textbf{92\% lower cost}. Each training stage contributes incrementally: SFT lifts the base model by 13.61 pp through high-quality trajectory imitation, while GRPO adds a further 6.63 pp by enabling the agent to discover novel strategies through online environmental interaction. A paired bootstrap test (10{,}000 resamples at the per-task level) yields a 95\% confidence interval of $[-1.27, 7.39]$ pp for the Gemini--GRPO pass@1 gap ($p > 0.05$). In contrast, the GRPO improvement over SFT ($+6.63$ pp, 95\% CI $[2.97, 9.64]$) is significant.

\textbf{Performance by Task Difficulty.} Table~\ref{tab:difficulty} breaks down pass@3 results by difficulty. GRPO improves over the base model on both standard (+13.16 pp) and hard tasks (+14.85 pp), with each training stage contributing: SFT provides the majority of the gain, while GRPO adds a further +2.63 pp on standard and +2.47 pp on hard tasks. On standard tasks, the gap to Gemini 3 Pro Preview is 3.95 pp; on hard tasks, the gap is 4.95 pp, indicating that the remaining frontier model advantage is relatively uniform across difficulty levels.

\section{Ablation \& Analysis}\label{ablation-analysis}

\subsection{Ablation Studies}\label{ablation-studies}

{\def\LTcaptype{none}
\begin{table}[t]
\caption{Success rate by task difficulty (pass@3).}
\label{tab:difficulty}
\small
\begin{tabular}{@{}lcc@{}}
\toprule
Model & Standard (76) & Hard (202) \\
\midrule
Qwen3-VL-32B (Base) & 71.05\% & 56.93\% \\
AliyunConsoleAgent-32B (SFT) & 81.58\% & 69.31\% \\
AliyunConsoleAgent-32B (SFT+GRPO) & 84.21\% & 71.78\% \\
Gemini 3 Pro Preview & 88.16\% & 76.73\% \\
\bottomrule
\end{tabular}
\end{table}
}

Table~\ref{tab:ablation} compares post-SFT training strategies on the 278-task benchmark (pass@1, Rule-based evaluation).

{\def\LTcaptype{none}
\begin{table}[t]
\caption{Training paradigm ablation (pass@1). Deltas are vs.\ SFT baseline.}
\label{tab:ablation}
\small
\begin{tabular}{@{}lcc@{}}
\toprule
Training Stage & pass@1 & $\Delta$ vs.\ SFT \\
\midrule
Qwen3-VL-32B (Base) & 43.28\% & --- \\
SFT (distillation) & 56.89\% & baseline \\
SFT $\rightarrow$ GRPO & 63.52\% & +6.63 pp \\
\bottomrule
\end{tabular}
\end{table}
}

SFT provides the largest absolute gain (+13.61 pp), establishing foundational console interaction capability. GRPO delivers a further +6.63 pp over SFT, and this improvement is statistically significant (95\% CI $[2.97, 9.64]$). GRPO's advantage stems from online exploration enabling the model to discover novel strategies through real-world interaction rather than imitating fixed demonstrations. GRPO training exhibits non-monotonic oscillations consistent with prior GUI agent RL work \cite{xu2025mobilerl,yang2025zerogui}; we use validation-based checkpoint selection every 5 steps, with the best checkpoint at step 25.

\subsection{Qualitative Analysis}\label{qualitative-analysis}

Analysis of 278 evaluation tasks reveals that GRPO introduces \textbf{no consistent regression}: no task exhibits a 3/3$\rightarrow$0/3 degradation pattern. Meanwhile, 9 tasks transition from consistent SFT failure to GRPO success ($\geq$2/3 pass). We identify two key capabilities acquired through RL:

\textbf{Precondition Construction.} The GRPO model exhibits human-like reasoning to establish prerequisite states. For instance, instructed to ``disable auto-renewal'' when the toggle is already OFF, the SFT model does nothing and fails. The GRPO model correctly deduces it must first \emph{enable} the feature to successfully perform the verifiable ``disable'' operation, demonstrating advanced goal-oriented reasoning.

\textbf{Adaptive Plan Adjustment.} The GRPO model develops the ability to deviate from document instructions when the prescribed UI path is unavailable. In a case involving RDS batch parameter modification, the document-specified ``Batch Modify'' button does not exist in the current UI. The SFT model enters an unrecoverable loop. The GRPO model, after detecting the discrepancy, autonomously switches to modifying parameters on individual instances---completing the task via an alternative path (2/3 success). This reflects a shift from \textbf{process-oriented} to \textbf{goal-oriented} execution.

\subsection{Error Analysis}\label{error-analysis}

Manual inspection of failed trajectories reveals three dominant failure modes. \textbf{(1)~Resource provisioning gaps} constitute the largest category: missing prerequisites, data not yet generated (e.g., backups requiring a 24-hour cycle), specification mismatches, and incorrect resource states. These are predominantly deterministic (fail in all 3 runs), reflecting Rollout Environment coverage gaps rather than agent errors; extending META templates with post-creation state conditioning would address the majority. \textbf{(2)~UI interaction failures} arise from frontend non-determinism: cascading dropdown race conditions, popup interference from platform AI assistants, greyed-out buttons due to hidden validation rules, and non-standard components (date pickers, file uploaders) incompatible with SoM-based interaction. Unlike resource failures, these are mostly intermittent (succeed in at least one run), suggesting that retry or wait-for-stability strategies could substantially reduce them. \textbf{(3)~Agent decision errors} reflect genuine reasoning gaps: premature task completion before the final confirmation step, wrong actions (parameter errors, reversed operations), and navigation failures. These represent the true capability ceiling and are the primary target for future RL iterations with denser per-step reward signals.

\subsection{Discussion}\label{discussion}

\textbf{Production Deployment and Impact.} The framework was deployed in Alibaba Cloud's pipeline (151 China-site, 123 international products) using frontier models (Jun 2025--Jan 2026), auditing 54,000+ procedures, identifying 4,399 confirmed defects shipped by product teams, and achieving a 91\% defect-confirmation rate and 76\% end-to-end success. However, frontier APIs incur prohibitive costs ($\sim$CNY 7/task) and privacy risks. As demonstrated in our evaluation (Section 6), AliyunConsoleAgent-32B achieves comparable performance on tasks drawn from these production documents, validating it as a ready-to-go, drop-in replacement. We are currently gray-testing the 32B model; at CNY 0.56/task, full adoption projects a $\sim$92\% ($\sim$CNY 350K) cost reduction, while the decoupled Rollout architecture supports $\sim$300 tasks/hour with 200+ concurrent sandboxes.

\textbf{Limitations.} The system faces three practical challenges deployed at scale. (1) \textbf{Provisioning Maintenance:} Our Rollout environment relies on static Terraform templates generated by frontier models to pre-provision required cloud resources. Because cloud platform APIs and documentation frequently update, these static templates become stale. Periodically regenerating these scripts via frontier APIs incurs massive ongoing costs, highlighting the need to train localized coding models or develop reusable resource templates. (2) \textbf{Visual Grounding:} Relying on Set-of-Mark annotations demands fragile DOM-parsing logic to extract frontend HTML elements. As multimodal models improve, future iterations should transition toward pure-vision interaction to bypass the maintenance burden of parsing complex, ever-changing frontend code. (3) \textbf{Infrastructure Expense:} The financial cost of running live verification is substantial. Testing documents that depend on premium resources (e.g., high-specification GPU instances) at the same frequency as basic services is economically unsustainable, requiring future deployments to dynamically adjust testing frequencies based on underlying infrastructure costs.

\section{Conclusion}\label{conclusion}

We present AliyunConsoleAgent, a web agent framework for real-world cloud console environments. Through two-stage training combining model trajectory distillation (SFT) with GRPO reinforcement learning using a dual-channel ORM, the model evolves from instruction following to autonomous decision-making. Our high-determinism Rollout system with Terraform pre-provisioning and ResourceCoder dynamic provisioning isolates environmental noise from the training signal. AliyunConsoleAgent-32B achieves 63.52\% mean success rate on 278 end-to-end cloud tasks---approaching Gemini 3 Pro Preview (bootstrap 95\% CI includes zero, $p > 0.05$)---at 92\% lower cost, making full-coverage periodic documentation verification feasible.

\newpage
\bibliographystyle{ACM-Reference-Format}
\bibliography{references}

\section*{GenAI Usage Disclosure}
In accordance with the ACM Policy on the use of Generative AI, we disclose that generative AI tools (Claude Code, Codex) were used to assist in polishing the English writing of this manuscript. The technical content, experimental design, and all reported results are entirely the work of the authors, who take full responsibility for the accuracy and integrity of the paper.

\end{document}